\documentclass[letterpaper]{article}
\usepackage{aaai}
\usepackage{times}

\usepackage[numbers]{natbib}
\usepackage{multicol}
\usepackage[bookmarks=true]{hyperref}

\usepackage{graphicx} 
\usepackage{booktabs}
\usepackage{amsmath} 
\usepackage{amssymb}  

\usepackage{algpseudocode}
\usepackage{algorithm}

\usepackage{subfigure}

\newcommand{\eqpunct}[1]{\text{#1}}

\DeclareMathOperator*{\argmax}{arg\,max}

\nocopyright

\begin{document}

\title{Planning for Decentralized Control of Multiple Robots Under Uncertainty}




%



\setlength\titlebox{2.5in}

\author{Christopher Amato$^{1}$ \and George D. Konidaris$^{1}$ \and Gabriel Cruz$^{1}$  \AND Christopher A. Maynor$^{2}$ \and Jonathan P. How$^{2}$ \and Leslie P. Kaelbling$^{1}$ \\  \\ $^{1}$CSAIL, $^{2}$LIDS \\MIT \\ Cambridge, MA 02139}


\maketitle

\begin{abstract}

 We describe a probabilistic framework for synthesizing control policies for general multi-robot systems, given environment and sensor models and a  cost function. Decentralized, partially observable Markov decision processes (Dec-POMDPs) are a general model of decision processes where a team of agents must cooperate to optimize some objective (specified by a shared reward or cost function) in the presence of uncertainty,  but where communication limitations mean that the agents cannot share their state, so execution must proceed in a decentralized fashion. While Dec-POMDPs are typically intractable to solve for real-world problems, recent research on the use of macro-actions in Dec-POMDPs has significantly increased the size of problem that can be practically solved as a Dec-POMDP. We describe this general model, and show how, in contrast to most existing methods that are specialized  to a particular problem class, it can synthesize control policies that use whatever opportunities for coordination are present in the problem, while balancing off uncertainty in outcomes, sensor information, and information about other agents. We use three variations on a warehouse task to show that a single planner of this type can generate cooperative  behavior using task allocation, direct communication, and signaling, as appropriate. 

\end{abstract}


\section{Introduction}


The decreasing cost and increasing sophistication of recently available robot hardware
has the potential to create many new opportunities for applications where teams of relatively cheap
robots can be deployed to solve real-world problems. Practical methods
 for coordinating such multi-robot teams are therefore becoming  critical. 
A wide range of approaches have been developed for solving specific classes of multi-robot problems, such as 
task allocation \cite{Gerkey04}, navigation in a formation \cite{Balch98}, cooperative transport of an object \cite{Kube00}, coordination with signaling \cite{beckers1994local} or communication under various limitations \cite{rekleitis2004limited}. Broadly speaking, the current
state of the art is to hand-design special-purpose controllers that are explicitly
designed to exploit some property of the environment or produce a specific desirable behavior. 
Just as in
the single-robot case, it would be much more desirable to instead specify a world model and a
cost metric, and then
have a general-purpose planner automatically derive a controller that minimizes cost, while
remaining robust to the uncertainty that is fundamental to real robot systems \cite{probrobot}.




The decentralized partially observable Markov decision process (Dec-POMDP) is a general
framework for representing multiagent coordination problems. 
Dec-POMDPs have been studied in fields such as control \cite{CDC13,Mahajan13}, operations research \cite{Bernstein02} and artificial intelligence  \cite{BernsteinUAI00,Oliehoek12RLBook}. 
Like the MDP \cite{Puterman94} and POMDP  \cite{Kaelbling98} models that it extends, the Dec-POMDP model is very general, considering uncertainty in outcomes, sensors and information about the other agents, and 
aims to optimize policies against a a general cost function. 
%
 Dec-POMDP problems are often characterized by incomplete or partial information about the environment and the state of other agents due to limited, costly or unavailable communication.
Any  problem where multiple agents share a single overall reward or cost function
can be formalized as a Dec-POMDP, which means a good Dec-POMDP solver would allow us to
automatically generate control policies (including policies over when and what to communicate)
for very rich decentralized control problems, in the presence of uncertainty. Unfortunately, this generality
comes at a cost: Dec-POMDPs are typically infeasible to solve except for very small problems \cite{AAMAS14AKK}.

One reason for the intractability of solving large Dec-POMDPs is that 
current approaches model problems at a low level of granularity, where each agent's actions are primitive operations lasting exactly one time step.  Recent research has
 addressed the more realistic  \textit{MacDec-POMDP} case where each agent has \emph{macro-actions}: temporally extended actions which may require different amounts of time to execute \cite{AAMAS14AKK}.
 This enables  systems to be modeled so that coordination decisions only occur at the level of deciding which macro-actions to execute. 
MacDec-POMDPs 
retain the ability to coordinate agents   while allowing near-optimal solutions to be generated for significantly larger problems than would be possible using other Dec-POMDP-based methods.


  Macro-actions
 are a natural model for the modular controllers often sequenced to obtain robot behavior. 
The macro-action approach leverages expert designed or learned controllers for solving subproblems (e.g., navigating to a waypoint or grasping an object), bridging the gap between traditional robotics research and work on Dec-POMDPs. 
This approach has the potential to produce high-quality general solutions for real-world heterogeneous multi-robot coordination problems by automatically generating control and communication policies, given a model.

We describe MacDec-POMDPs
and argue for their use as a general model for multi-robot systems. We begin by formally describing the Dec-POMDP model, its solution and relevant properties, and then extend it the definition to include MacDec-POMDPs. 
We describe an approximate, memory bounded algorithm for solving MacDec-POMDPs, and a process
for converting a robot domain into a MacDec-POMDP model, solving it, and using the solution to produce
 a SMACH \cite{Bohren} finite-state machine task controller.
Finally, we use three variations of a 
 warehouse task to show that a MacDec-POMDP planner
  can generate cooperative  behavior that allocates tasks, uses direct communication,
 and employs signaling, as appropriate. The appropriate coordination behaviors emerge as properties
 of the optimal solution for each individual model. 




\section{Decentralized, Partially Observable Markov Decision Processes }

Dec-POMDPs \cite{Bernstein02} generalize partially observable Markov decision processes to the multiagent,
 decentralized setting.
Multiple agents  operate under uncertainty based on (possibly different) partial views of the world, with execution  unfolding over a bounded or unbounded sequence of steps. At each step, every agent chooses an action (in parallel) based purely on locally observable information, resulting in an immediate reward and an observation being obtained 
by each individual agent. The agents share a single reward or cost
function, so they should cooperate to solve the task, but their local views mean that operation is decentralized
during execution. 

\begin{figure}[ht!!]
      \centering
       \includegraphics[width=.95\columnwidth]{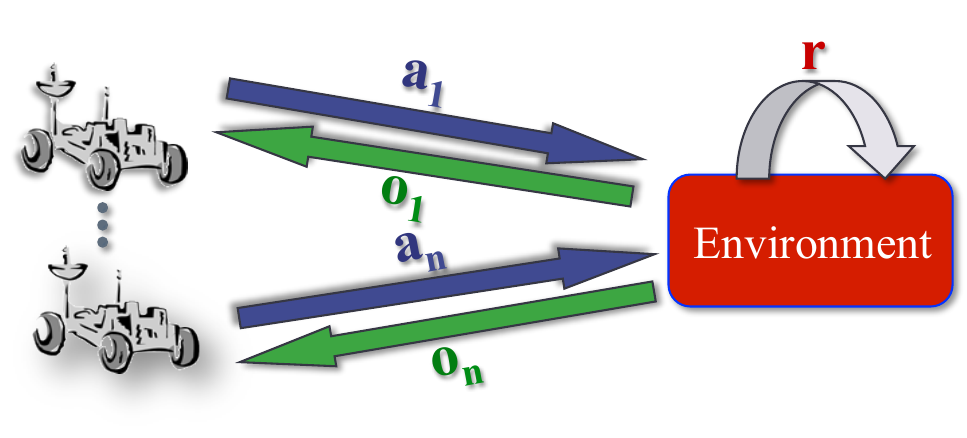}
      \caption{Representation of $n$ agents in a Dec-POMDP setting with actions $a_i$  and observations $o_i$ for each agent $i$ along with a single reward $r$.}
      \label{fig:DecPOMDP}
   \end{figure}

As depicted in Fig.~\ref{fig:DecPOMDP}, a Dec-POMDP \cite{Bernstein02} involves multiple agents that operate under uncertainty based on different streams of observations. We focus on solving sequential decision-making problems with discrete time steps and stochastic models with finite states, actions, and observations, though the model can be extended to continuous problems. A key assumption is that state transitions are \emph{Markovian}, meaning that the state at time $t$ depends only on the state and events at time $t - 1$.
The reward is typically only used as a way to specify the objective of the problem and is not observed during execution. 

%


More formally, a  Dec-POMDP is described by a tuple $ \langle  I, S, \{A_i\}, T, R, \{\Omega_i\}, O, h\rangle $,
where
\begin{itemize}
\item $I$ is a finite set of agents.
\item $S$ is a finite set of states with designated initial state
distribution $b_0$.
\item  $A_i$ is a finite set of actions for each agent $i$ with $A=\times_i A_i$ the set of joint actions, where $\times$ is the Cartesian product operator.
\item  $T$ is a state transition probability function, $T: S \times A \times S \to [0,1]$, that specifies the
probability of transitioning from state $s \in S$ to $s' \in S$ when the actions $\vec{a} \in A$ are taken by the agents. Hence, $T(s,\vec{a},s') = \Pr(s' | \vec{a},s)$.
\item  $R$ is a reward function: $R: S \times A \to \mathbb{R}$, the immediate reward for being in state $s \in S$ and taking the
actions $\vec{a} \in A$.
\item $\Omega_i$ is a finite set of observations for each agent, $i$, with $\Omega=\times_i \Omega_i$ the set of joint observations.
\item  $O$ is an observation probability function: $O: \Omega \times A \times S \to [0,1]$, the
probability of  seeing  observations $\vec{o} \in \Omega$  given 
actions $\vec{a} \in A$  were taken which results in state $s' \in S$. Hence $O(\vec{o},\vec{a},s') = \Pr(\vec{o} | \vec{a},s')$.
\item $h$ is the number of steps until the problem terminates, called the horizon.
\end{itemize}


Note that while the actions and observation are factored, the state need not be. This flat state representation allows more general state spaces with arbitrary state information outside of an agent (such as target information or environmental conditions). 
Because the full state is not directly observed, it may be beneficial for each agent to remember a history of its observations. Specifically, we can consider an action-observation history for agent $i$ as $$H^A_i=(s_i^0, a_i^1,\ldots,s_i^{l-1},a_i^l).$$ 
Unlike in POMDPs, it is not typically possible to calculate a centralized estimate of the system state from the observation history of a single agent, because the system state depends on the behavior of all of the agents. 


\subsection{Solutions}

\begin{figure}[ht!!!]
      \centering
\subfigure[]{\includegraphics[width=.6\hsize]{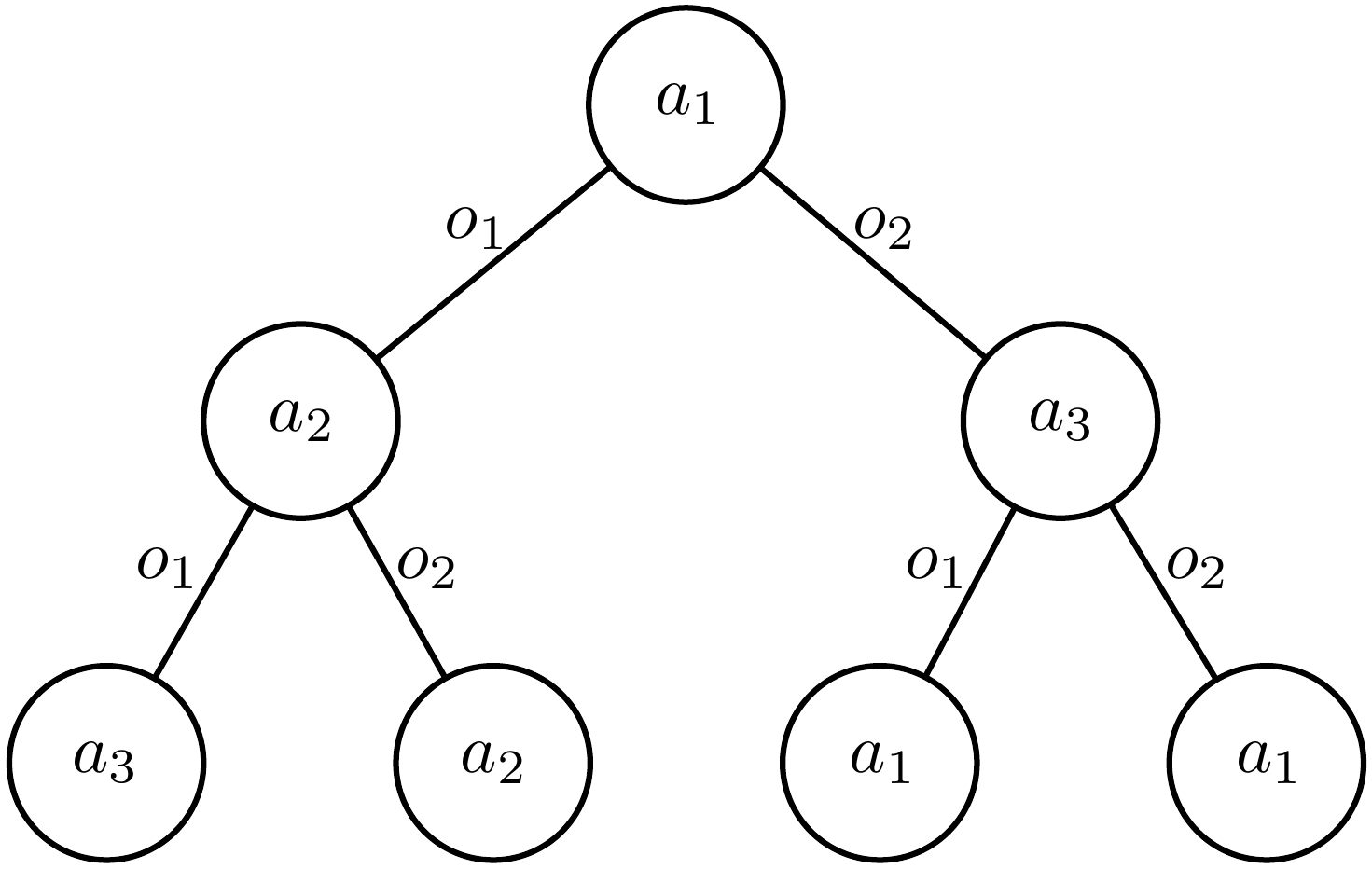}}
\quad
\subfigure[]{\includegraphics[width=.3\hsize]{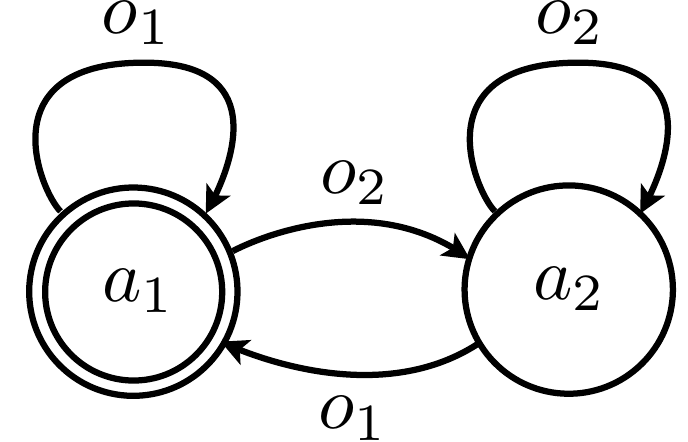}}
      \caption{A single agent's policy represented as (a) a policy tree and (b) a finite-state controller with initial state shown with a double circle.}
      \label{fig:TreeAndCont}
   \end{figure}

A solution to a Dec-POMDP is a \emph{joint policy}---a set of policies, one for each agent in the problem. 
Since each policy
is a function of history, rather than of a directly observed state, it is typically represented as either a policy tree,
where the vertices  indicate actions to execute and the edges indicate transitions conditioned on an observation,
or as a finite state controller which executes in a similar manner. An example of each is given in Figure \ref{fig:TreeAndCont}.

As in the POMDP case, the goal is to maximize the total cumulative reward, beginning at some initial distribution over states $b_0$.  In general, the agents do not observe the actions or observations of the other agents, but the rewards, transitions, and observations depend on the decisions of all agents. The work discussed in this paper (and the vast majority of work in the Dec-POMDP community) considers the case where the model is assumed to be known to all agents.

The value of a joint policy, $\pi$, from state $s$ is
$$ V^\pi(s)=\mathop{\rm E} \left[ \sum_{t=0}^{h-1} \gamma^tR(\vec a^t, s^t) | s,\pi \right]\eqpunct{,}$$
which represents the expected value of the immediate reward for the set of agents summed for each step of the problem given the action prescribed by the policy until the horizon is reached. In the finite-horizon case,  the discount factor, $\gamma$, is typically set to 1. In the infinite-horizon case, as the number of steps is infinite, the discount factor $\gamma \in [0,1)$ is included to maintain a finite sum and $h=\infty$. An  \emph{optimal policy} beginning at state $s$ is 
$\pi^*(s) = \argmax_{\pi} V^{\pi}(s)$.


%
Unfortunately, large problem instances remain intractable: some advances have been made in optimal algorithms \cite{CDC13,ICAPS09,Aras07,Boularias08,IJCAI13,JAIR13},
but optimally solving a Dec-POMDP  is NEXP-complete, so most approaches that scale well 
  make very strong assumptions about the domain (e.g., assuming a large amount of independence between 
  agents) \cite{AAMAS13,Melo2011,Nair05} and/or have no guarantees about solution quality \cite{Oliehoek13AAMAS,Sven07a,Pras11}. 

\section{Macro-Actions for Dec-POMDPs}

Dec-POMDPs typically require synchronous decision-making: every agent repeatedly determines which action
to execute, and then executes it within a single time step.  This restriction is problematic for robot domains
for two reasons. First, robot systems are typically endowed with a set of controllers, and planning consists of
sequencing the execution of those controllers. However, due to both environmental and controller complexity, the
controllers will almost always execute for an extended period, and take differing amounts of time to run. 
Synchronous decision-making would thus require us to wait until all robots have completed their controller execution
before we perform the next action selection, which is suboptimal and may not even always be possible (since the robots
do not know the system state and staying in place may be difficult in some domains). Second,
the planning complexity of a Dec-POMDP is doubly exponential in the horizon.
 A planner that must try to reason about
all of the robots' possible policies at every time step will only ever be able to make very short plans. 

Recent research has extended the Dec-POMDP  model 
to plan using \textit{options}, or temporally extended actions \cite{AAMAS14AKK}. This MacDec-POMDP formulation models
a group of robots that must plan by sequencing an existing set of controllers,
enabling planning at the appropriate level to  compute near-optimal solutions for problems with significantly longer 
horizons and larger state-spaces.


We can gain additional benefits by
exploiting known structure in the multi-robot problem.  
For instance, most controllers
only depend on locally observable information and do not require coordination. 
For example, consider a controller that navigates a robot to a waypoint. Only local information is required for 
navigation---the robot may detect other robots but their presence does not change its objective, and it 
simply moves around them---but 
choosing the target waypoint likely requires the planner to consider the locations and actions of all robots.  
Macro-actions with independent execution allow coordination decisions to be made only when  necessary (i.e., when choosing macro-actions) rather than at every time step.  
Because we build on top of Dec-POMDPs, macro-action choice may depend on history, but during execution macro-actions may depend only on a single observation, depend on any number of steps of history, or even represent the actions of a set of robots. That is, macro-actions are very general and can be defined in such a way to take advantage of the knowledge available to the robots during execution. 


%



\subsection{Model}

We first consider macro-actions that only depend on a single robot's information. This is an extension the \emph{options framework} \cite{Sutton99}| to multi-agent domains while dealing with the lack of synchronization between agents.  
The options framework is a formal model of a macro-actions \cite{Sutton99} that has been very successful in aiding representation and solutions in single robot domains  \cite{Kober13}.
A   MacDec-POMDP with local options is defined as a  Dec-POMDP where we also assume $M_i$ represents a finite set of options for each agent, $i$, with $M=\times_i M_i$ the set of joint options \cite{AAMAS14AKK}. A  \emph{local option} is defined by the tuple:
$$M_i=(\beta_{m_i}, \mathcal{I}_{m_i}, \pi_{m_i}),$$ 
consisting of stochastic termination condition $\beta_{m_i}: H^A_i \to [0,1]$, initiation set $\mathcal{I}_{m_i} \subset H^A_i$ and option policy $\pi_{m_i}: H^A_i\times  A_i \to [0,1]$. 
Note that this representation uses action-observation histories of an agent in the terminal and initiation conditions as well as the option policy.  
Simpler cases can consider reactive policies that map single observations to actions as well as termination and initiation sets that depend only on single observations. This is especially appropriate when the agent has knowledge about aspects of the state necessary for option execution (e.g., its own location when navigating to a waypoint causing observations to be location estimates).  
As we later discuss, initiation and terminal conditions can depend on global states (e.g., also ending execution based on unobserved events). 

Because it may be beneficial for agents to remember their histories when choosing which
option to execute, we consider policies that remember option histories (as opposed to action-observation histories). We define an \emph{option history} as $$H^M_i=(h_i^0, m_i^1,\ldots,h_i^{l-1},m_i^l)$$ which includes both the action-observation histories where an option was chosen and the selected options themselves. 
The option history also provides a nice representation for using histories within options. 
It is more natural for option policies and termination conditions to depend on histories that begin when the option is first executed (action-observation histories) while the initiation conditions would depend on the histories of options already taken and their results (option histories). 
While a history over primitive actions also provides the number of steps that have been executed in the problem (because it includes actions and observations at each step), an option history may require many more steps to execute than the number of options listed. 
We can also define a (stochastic) local policy, $\mu_i: H^M_i \times M_i \to [0,1]$ that depends on option histories. We then define a joint policy for all agents as $\mu$.

Because option policies are built out of primitive actions, we can evaluate policies in a similar way to other Dec-POMDP-based approaches. 
Given a joint policy, the primitive action at each step is determined by the high level policy which chooses the option and the option policy which chooses the action. 
The joint policy and option policies can then be evaluated as:
$$ V^\mu(s)=\mathop{\rm E} \left[ \sum_{t=0}^{h-1} \gamma^tR(\vec a^t, s^t) | s,\pi,\mu \right].$$
For evaluation in the case where we define a set of options which use observations (rather than histories) for initiation, termination and option policies (while still using option histories to choose options) see 
Amato, Konidaris and Kaelbling \cite{AAMAS14AKK}.

\subsection{Algorithms}

The goal of MacDec-POMDP planning is to obtain 
a \emph{hierarchically optimal policy}: $\mu^*(s) = \text{argmax}_{\mu} V^{\mu}(s)$. This policy 
is the one that obtains the highest expected value
by sequencing the agent's given options. 
This policy may have a lower value than the optimal policy for the Dec-POMDP, because it does not
include all possible history-dependent low-level policies---the policies are restricted to be sequences
of macro-actions. We can guarantee that a globally optimal policy will be found by
 including the primitive actions in the set of options for each agent, but it
 typically makes little sense to do so, especially when our macro-actions model existing
 robot controllers. 

Because Dec-POMDP algorithms produce policies mapping agent histories to actions, they can be extended to consider options instead of primitive actions. We discuss how options can be incorporated into two such algorithms; extensions can also be made to other approaches. 

In both cases, deterministic polices are generated which are represented as policy trees (as shown in Figure \ref{fig:TreeAndCont}). A policy tree for each agent defines a policy that can be executed based on local information. The root node defines the option to choose in the known initial state, and another
option is assigned to  each of the legal terminal states of that option; this continues for the depth of the tree. Such a tree can be evaluated up to a desired (low-level) horizon using the policy evaluation given above, which may not reach some nodes of the tree due to the differing execution times of some options. 





\subsubsection{Dynamic Programming\\}

\begin{figure}
\subfigure[Step 1]{\includegraphics[width=.5in]{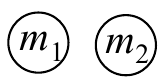}} \hspace{10pt}
\subfigure[Step 2 of DP]{\includegraphics[width=2.6in]{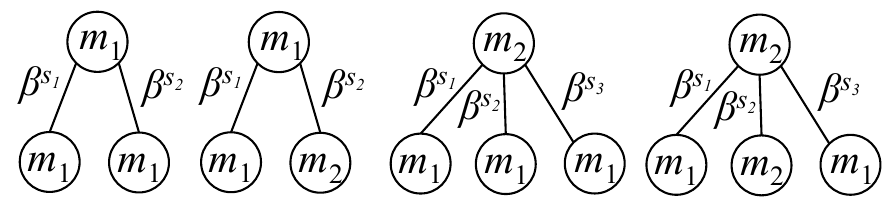}}
\caption{Policies for a single agent after one step of dynamic programming using options $m_1$ and $m_2$ where (deterministic) terminal states for options are represented as $\beta^s$.}
\label{fig:Trees} 
\end{figure}

A simple exhaustive search method can be used to generate hierarchically optimal deterministic policies which utilize options. This algorithm is similar in concept to the dynamic programming algorithm used in Dec-POMDPs \cite{Hansen04}, but full evaluation and pruning (removing dominated policies) are not used at each step. Instead we can exploit the structure of options to reduce the space of policies considered.

 We can exhaustively generate all combinations of options by first considering each agent using any one of it's options to solve the problem, as seen for one agent in Figure \ref{fig:Trees}. We can test all combinations of these 1-option policies for the set of agents to see if they are guaranteed to reach horizon $h$ (starting from the initial state). If any combination of policies does not reach $h$ with probability 1, an exhaustive backup is performed by considering starting from all possible options and then for any terminal condition of the option (represented as local terminal states $\beta^s$ in the figure), transitioning to one of the 1-option policies. This step creates all possible next (option) step policies.  
  We can check again to see if any of the current set of policies will terminate before the desired horizon and continue to grow the policies as necessary. When all policies are sufficiently long, all combinations of these  policies can be evaluated as above (by flattening out the  polices into primitive action Dec-POMDP policies, starting from some initial state and proceeding until $h$). The combination with the highest value at the initial state, $s_0$, is chosen as the option policy. Pseudocode for this approach is given in Algorithm \ref{alg:DP}.
 
 \begin{algorithm}
\caption{\label{alg:DP}Option-based dynamic programming (O-DP)}
\begin{algorithmic}[1]
\Function{OptionDecDP}{$h$}
\State $t \gets 0$
\State $someTooShort \gets true$
\State $\mu_{t} \gets \emptyset $
\Repeat
    \State $\mu_{t+1} \gets $ExhaustiveBackup$(\mu_{t})$
    \State $someTooShort \gets $TestPolicySetsLength($\mu_{t+1}$)
	\State $t \gets t + 1$
\Until{$someTooShort =false$}
\State Compute $V^{\mu_{t}}(s_0)$
\State \Return $\mu_t$
\EndFunction
\end{algorithmic}
\end{algorithm}

 This algorithm will  produce a hierarchically optimal deterministic policy  because it  constructs all legal deterministic option policies that are guaranteed to reach horizon $h$. This follows from the fact that options must last at least one step and all combinations of options are generated at each step until it can be guaranteed that additional backups will cause redundant policies to be generated. 
 Our approach represents exhaustive search in the space of legal policies that reach a desired horizon. As such it is not a true dynamic programming algorithm, but additional ideas from dynamic programming for Dec-POMDPs \cite{Hansen04} can be incorporated. For instance, we could prune policies based on value, but this method requires evaluating all possible joint policies at every state after each backup. This would require flattening the policy after each backup and evaluating all combinations of flat policies for all states starting from all possible reachable horizons. Instead, the benefit of this approach is that only legal policies are generated using the initiation and terminal conditions for options. As seen in Figure \ref{fig:Trees}, option $m_1$ has two possible terminal states while option $m_2$ has three. Furthermore, only option $m_1$ is applicable in local states $s_1$ and $s_3$. This structure limits the branching factor of the policy trees produced and thus the number of trees to be considered.

\subsubsection{Memory-Bounded Dynamic Programming\\}

Memory-bounded dynamic programming  (MBDP) \cite{Sven07a} can also be extended to use options as shown in Algorithm \ref{alg:MBDP}.  
Here, only a finite number of policy trees are retained (given by parameter \emph{MaxTrees}) after each backup. 
After an exhaustive backup has been performed, a set of $t$-step trees is chosen by evaluating the trees at states that are generated by a heuristic policy ($H_{pol}$ in the algorithm) that is executed for the first $h-t-1$ steps of the problem. A set of $MaxTrees$ states is generated and the highest valued trees for each state are kept. This process of exhaustive backups and retaining $MaxTrees$  trees continues, using shorter and shorter heuristic policies until the all combinations of the retained trees reach horizon $h$. Again, the set of trees with the highest value at the initial state is returned.

\begin{algorithm}[ht]
\caption{Option-based memory bounded dynamic programming (O-MBDP)}
\label{alg:MBDP}
\begin{algorithmic}[1]
\Function{OptionMBDP}{$MaxTrees$,$h$,$H_{pol}$}
\State $t \gets 0$
\State $someTooShort \gets true$
\State $\mu_{t} \gets \emptyset $
\Repeat
    \State $\mu_{t+1} \gets $ExhaustiveBackup$(\mu_{t})$
    \State Compute $V^{\mu_{t+1}}$
    \State  $\hat \mu_{t+1}  \gets \emptyset$
       \ForAll{$k \in MaxTrees$} 
     \State  $s_k \gets$ GenerateState($H_{pol}$,$h-t-1$)
   \State $\hat \mu_{t+1} \gets  \hat \mu_{t+1} \cup  \arg\max_{\mu_{t+1}} V^{\mu_{t+1}}(s_k)$
     \EndFor
	\State $t \gets t + 1$
	\State $ \mu_{t+1}  \gets \hat \mu_{t+1}$
\Until{$someTooShort =false$}
\State \Return $\mu_t$
\EndFunction
\end{algorithmic}
\end{algorithm}

This approach is potentially suboptimal because 
a fixed number of trees are retained, and tree sets are optimized over states that are both known and may never be reached. Nevertheless, since the number of policies retained at each step is bounded by $MaxTrees$, MBDP has time and space complexity linear in the horizon. As a result, MBDP (and its extensions \cite{ICAPS09,AkshatAAMAS10,Wu10}) have been shown to work well in many large Dec-POMDPs.  The option-based extension of MBDP uses the structure provided by the initiation and terminal conditions just like the dynamic programming approach in Algorithm \ref{alg:DP}, but does not have to produce all policies that will reach horizon $h$. Scalability can therefore be adjusted by reducing the $MaxTrees$ parameter (although solution quality may be reduced for smaller $MaxTrees$).





\section{Planning using MacDec-POMDPs in the Warehouse Domain}

\begin{figure}[!!!ht]
\includegraphics[width=3.4in]{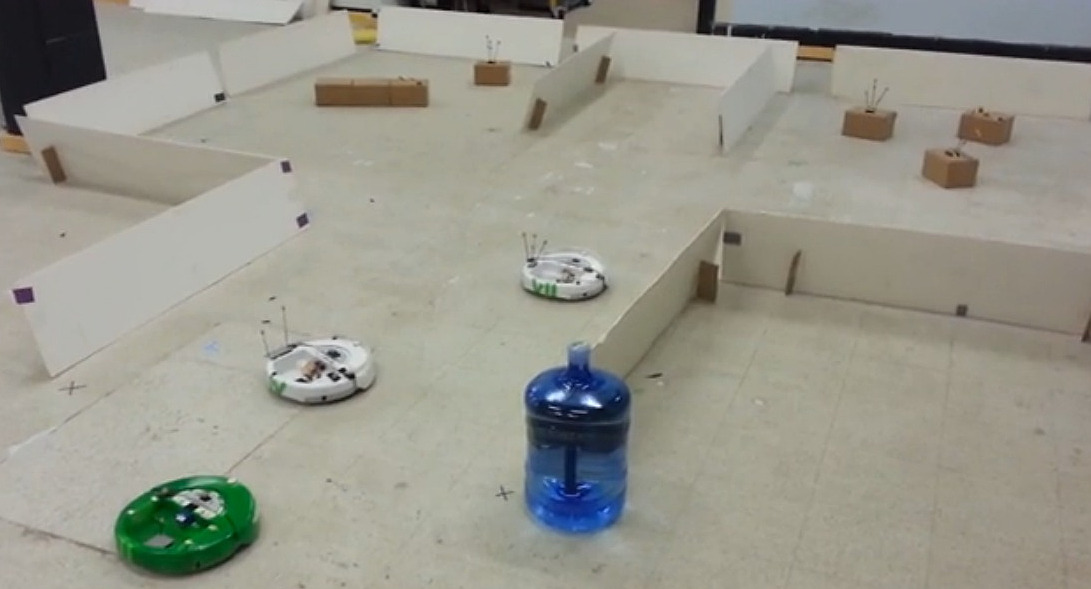}
\caption{The warehouse domain with three robots.}
\label{fig:warehouse} 
\end{figure}

We test our methods in a warehousing scenario using a set of iRobot Creates (Figure \ref{fig:warehouse}). We demonstrate 
how the same general model and solution methods can be applied in versions of this domain with different 
communication capabilities. 



\subsection{The Warehouse Domain}

We consider three robots in a warehouse tasked
with retrieving two different sized boxes: large and small. Robots can navigate to known depot locations (rooms) to retrieve boxes and bring them back to a designated drop-off area. The larger boxes require two robots to move (if a robot tries to pickup the large box by itself, it will move to the box, but fail to pick it up). 
While the locations of the depots are known, the contents (the number and type of boxes) are unknown. The solution for these problems is generated offline using our planner to produce a SMACH controller for each of the robots which are then executed  online in a decentralized manner. 


In each case we assumed that the robots could observe their own location (in a discretized version of the space), see other robots if they were within (approximately) one meter, observe the nearest box when in a depot and observe the size of the box if it is holding one.  
The resulting state space of the problem includes the location of the robots (discretized into nine possible locations) and the location of each of the boxes (in a particular depot, with a particular robot or at the goal). The three robot version of this scenario has 1,259,712,000 states, which is several orders of magnitude larger than problems typically solvable by Dec-POMDP solvers. These problems are solved using the option-based MBDP algorithm initialized with a hand coded heuristic policy.

Navigation was assumed to have a small amount of noise (reflecting the nonholonomic dynamics) in the amount of time required to move to locations, and this noise increases when the robots were pushing the large box (reflecting the need for slower movements and turns in this case). These noise parameters were assumed to be known in this work, but they could also be learned by executing macro-actions multiple times in the given initiation sets. 
We defined macro-actions that depend only on these observations, but again choosing which option to execute depends on the history of options executed and observations seen as a result (the option history). 


\subsection{Scenario 1: No Communication}

In the first scenario, we consider the case where robots could not communicate with each other. Therefore, all cooperation is based on the controllers that are generated by the planner (which knows the controllers generated for all robots when planning offline) and observations of the other robots (when executing online). The macro-actions were as follows:
\begin{itemize}
	\item Go to depot 1.
	\item Go to depot 2.
	\item Go to the drop-off area.
	\item Pick up the small box.
	\item Pick up the large box.
	\item Drop off a box.
\end{itemize}

The depot macro-actions are applicable anywhere and terminate when the robot is within the walls of the appropriate depot. The drop-off and drop macro-actions are only applicable if the robot is holding a box, and the pickup macro-actions are only applicable when the robot observes a box of the particular type. Navigation is stochastic in the amount of time that will be required to succeed (as mentioned above). Picking up the small box was assumed to succeed deterministically, but this easily be changed if the pickup mechanism is less robust. 
These macro-actions correspond to natural choices for robot controllers.

\begin{figure*}
\subfigure[Two robots set out for different depots.]{\includegraphics[width=1.6in]{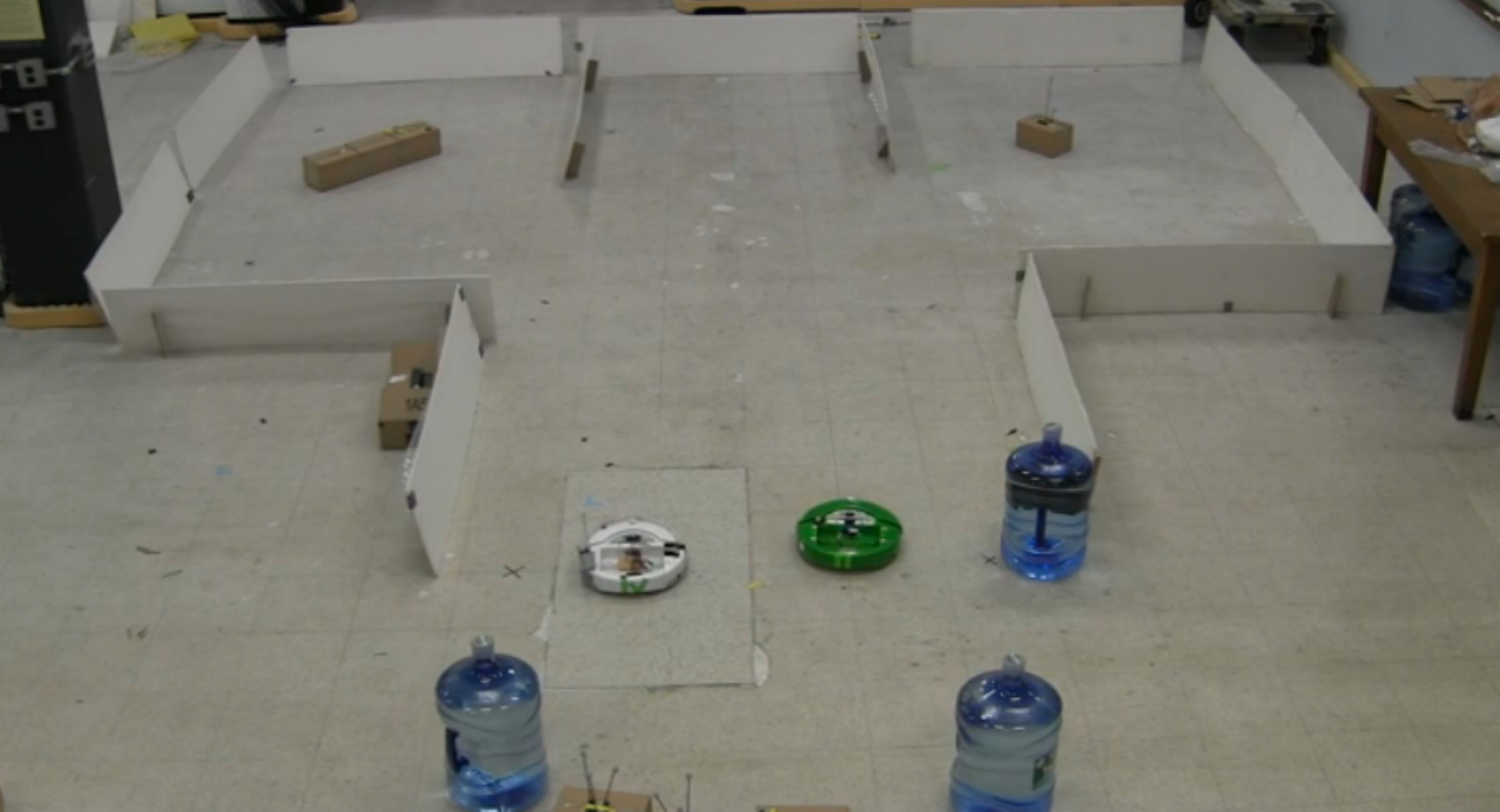}\label{Na}}  \hfill
\subfigure[The robots observe the boxes in their depots (large on left, small on right).]{\includegraphics[width=1.6in]{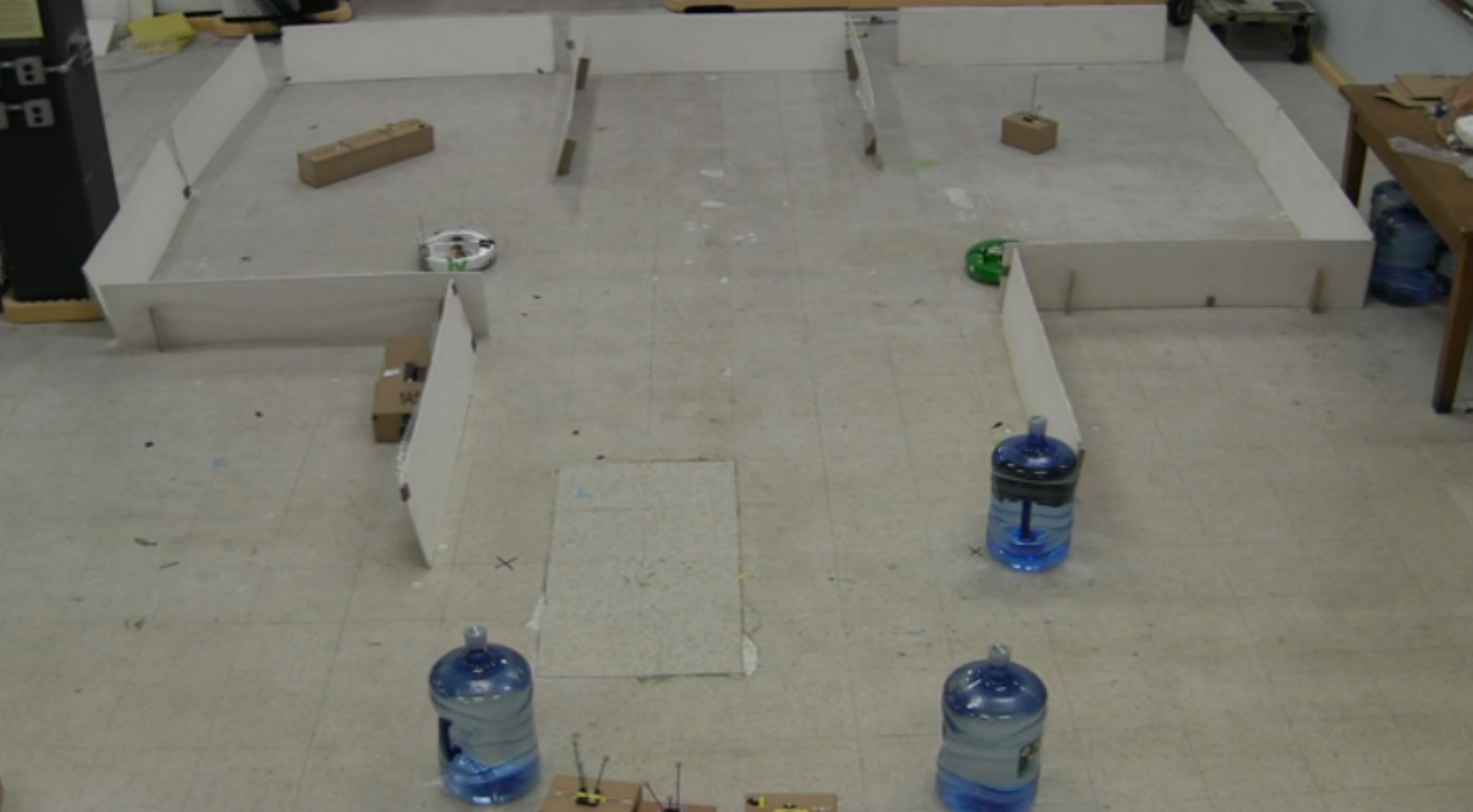}\label{Nb}}  \hfill
\subfigure[White robot moves to the large box and green robot moves to the small one.]{\includegraphics[width=1.6in]{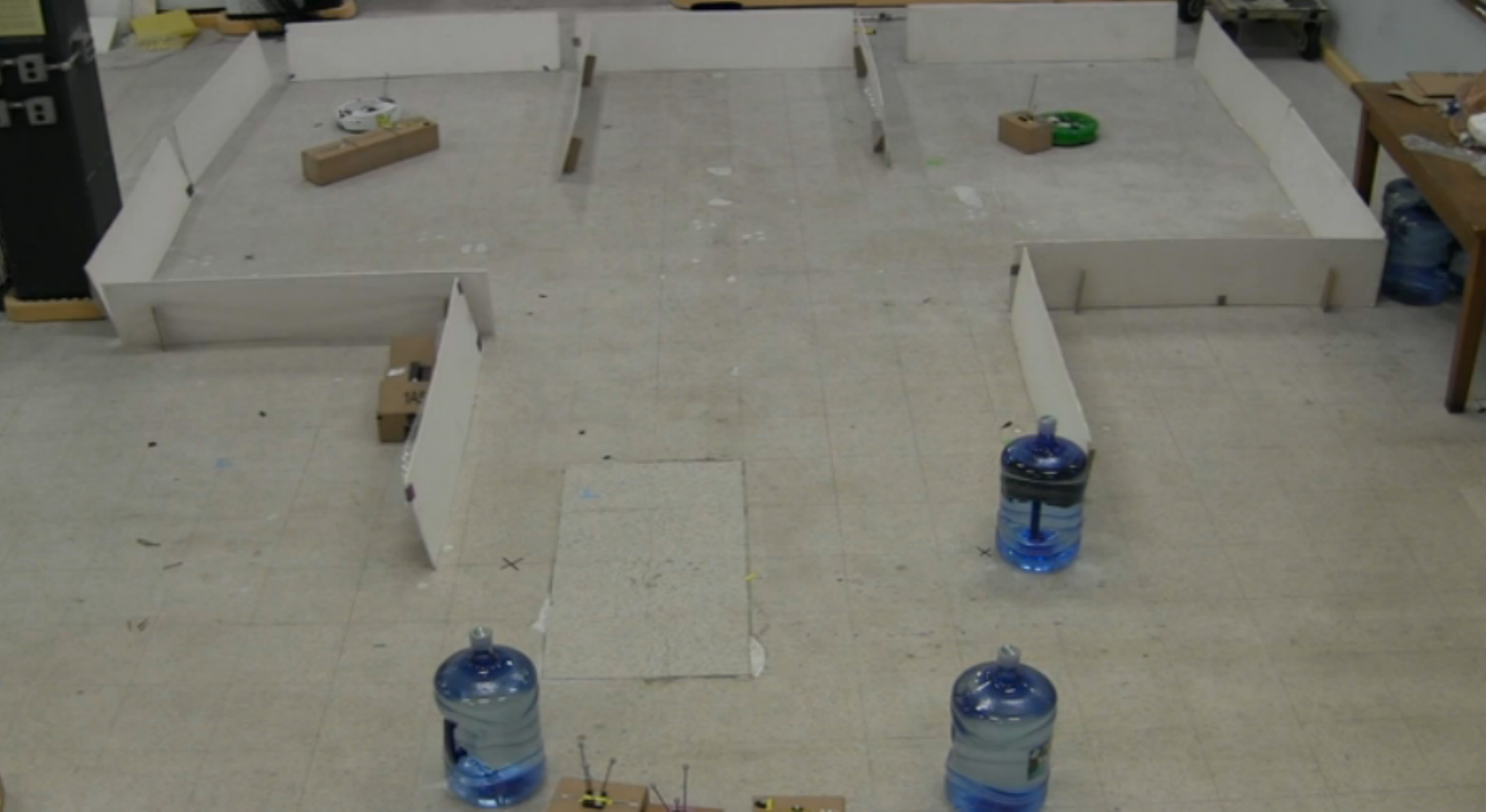}\label{Nc}} \hfill 
\subfigure[White robot waits while green robot pushes the small box.]{\includegraphics[width=1.6in]{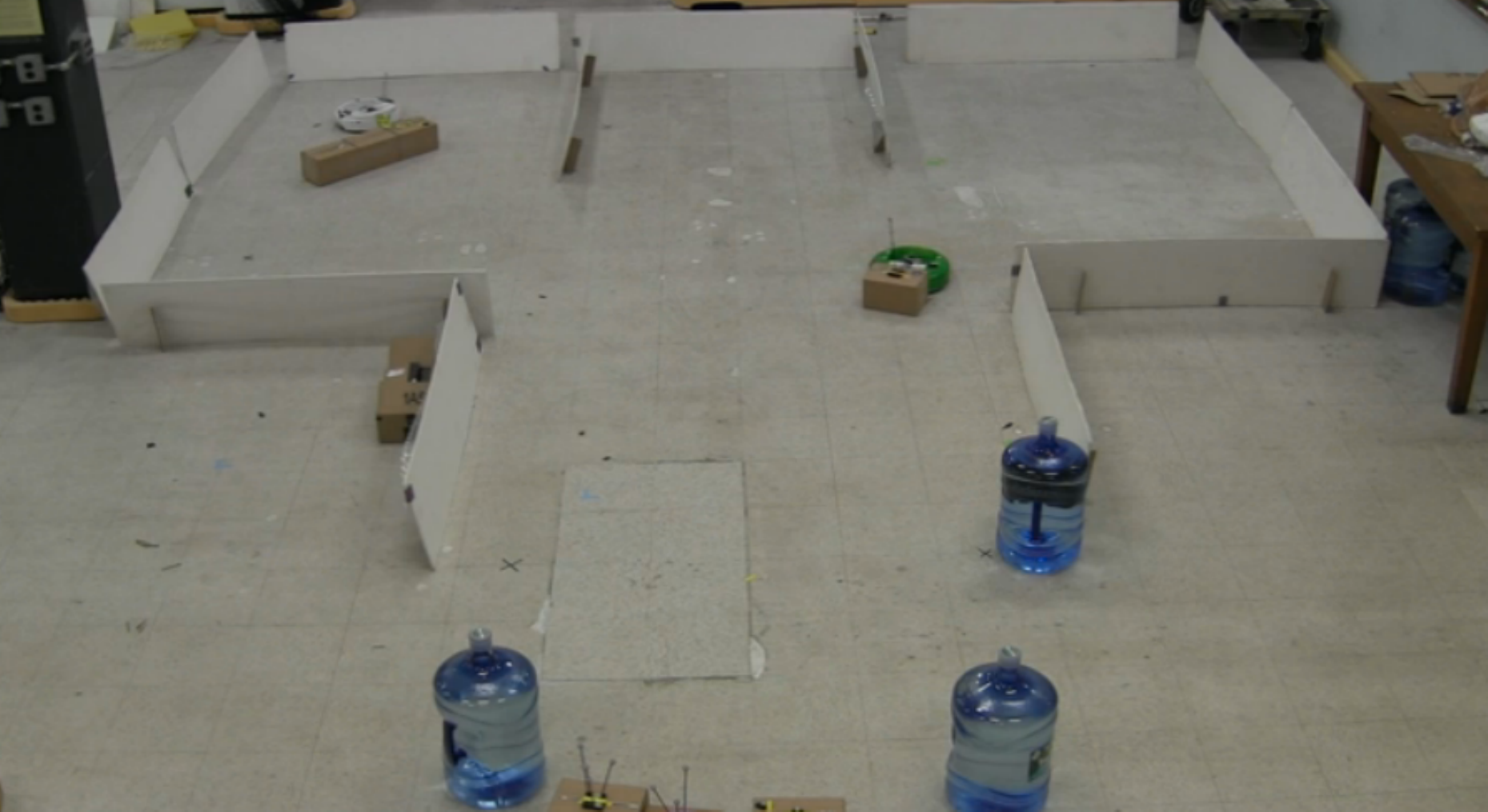}\label{Nd}}  \hfill
\subfigure[Green robot drops the box off at the goal.]{\includegraphics[width=1.6in]{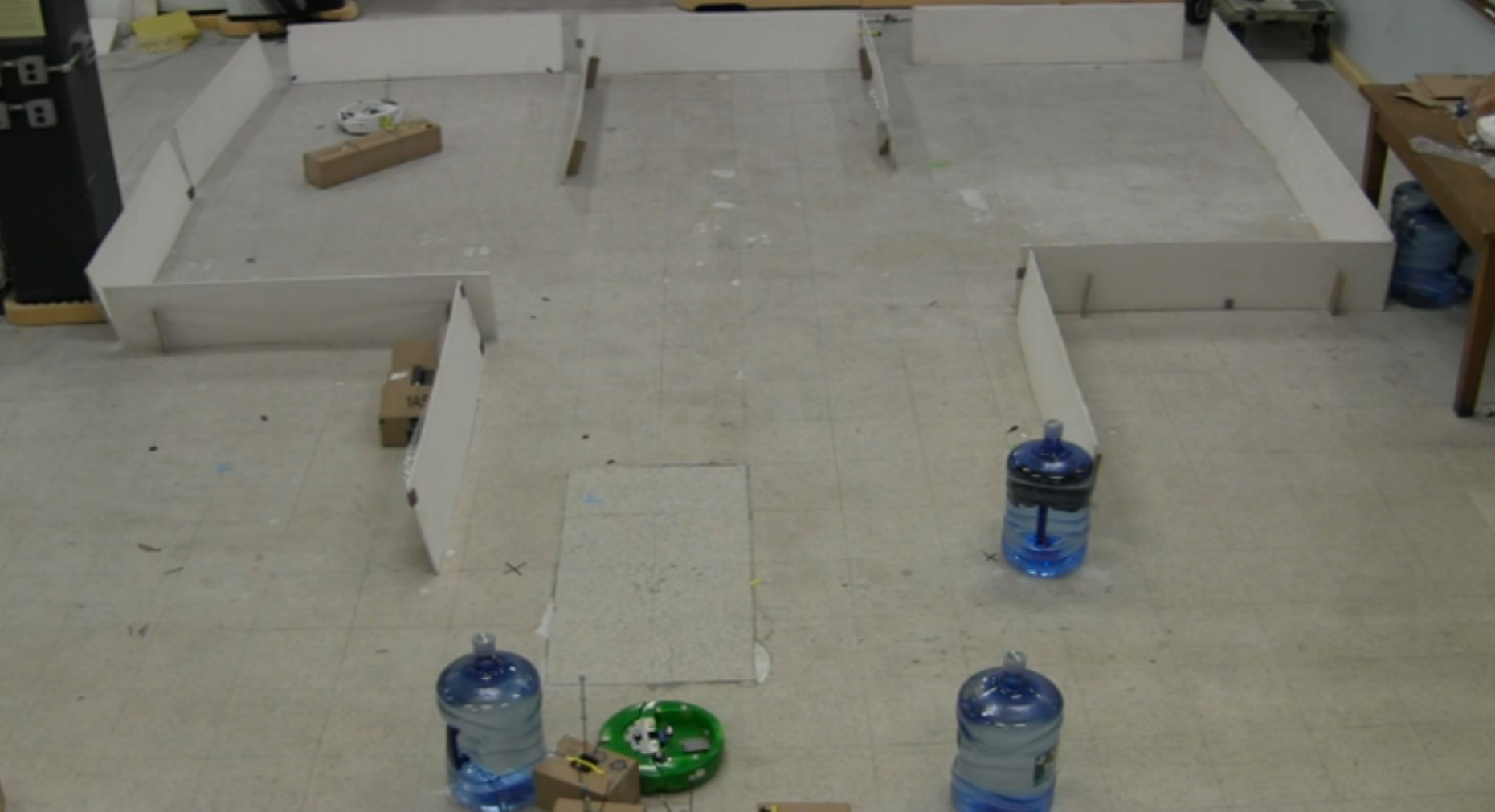}\label{Ne}}  \hfill
\subfigure[Green robot goes to the depot 1 and sees the other robot and large box.]{\includegraphics[width=1.6in]{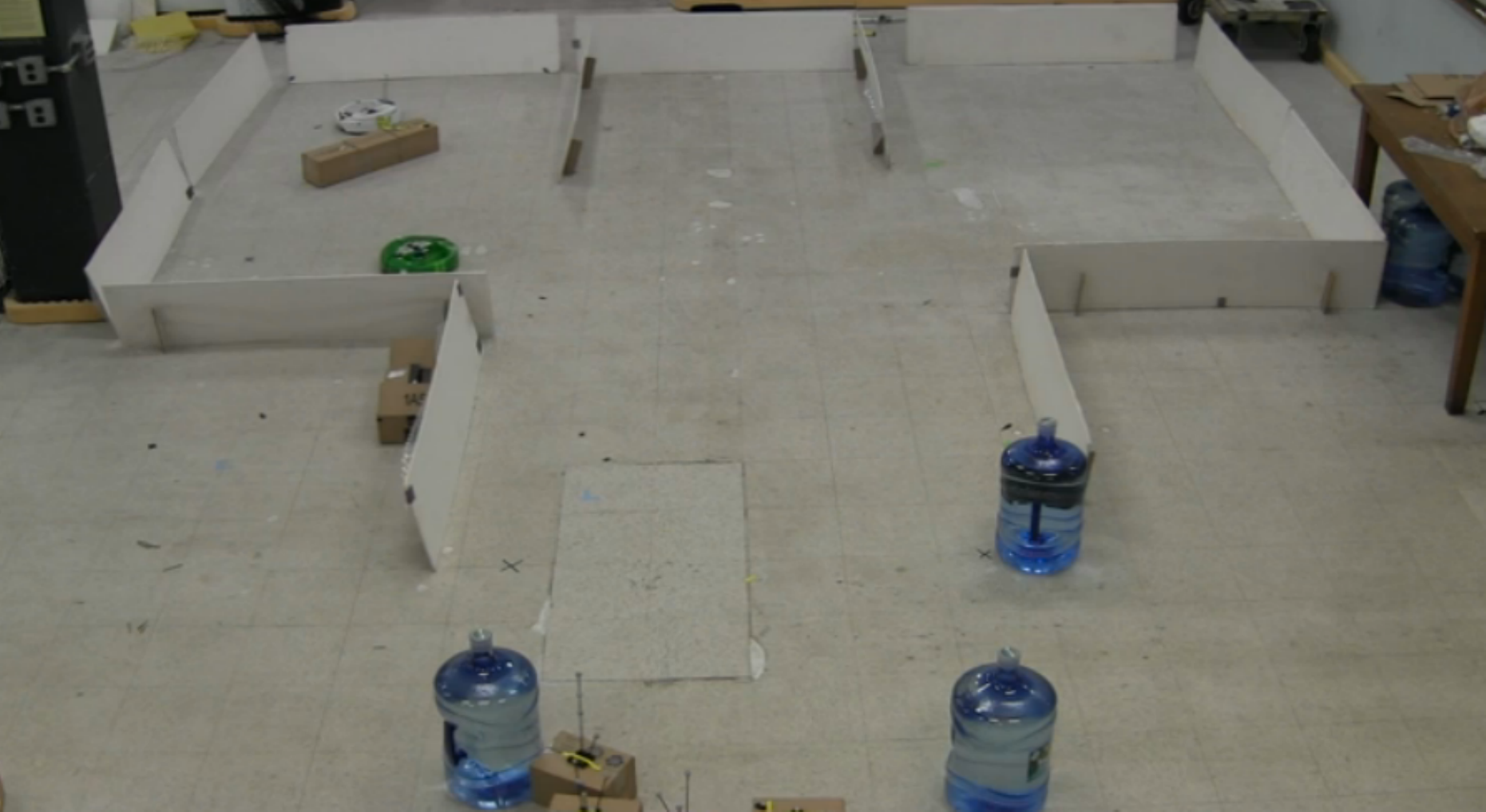}\label{Nf}}  \hfill
\subfigure[Green robot moves to help the white robot.]{\includegraphics[width=1.6in]{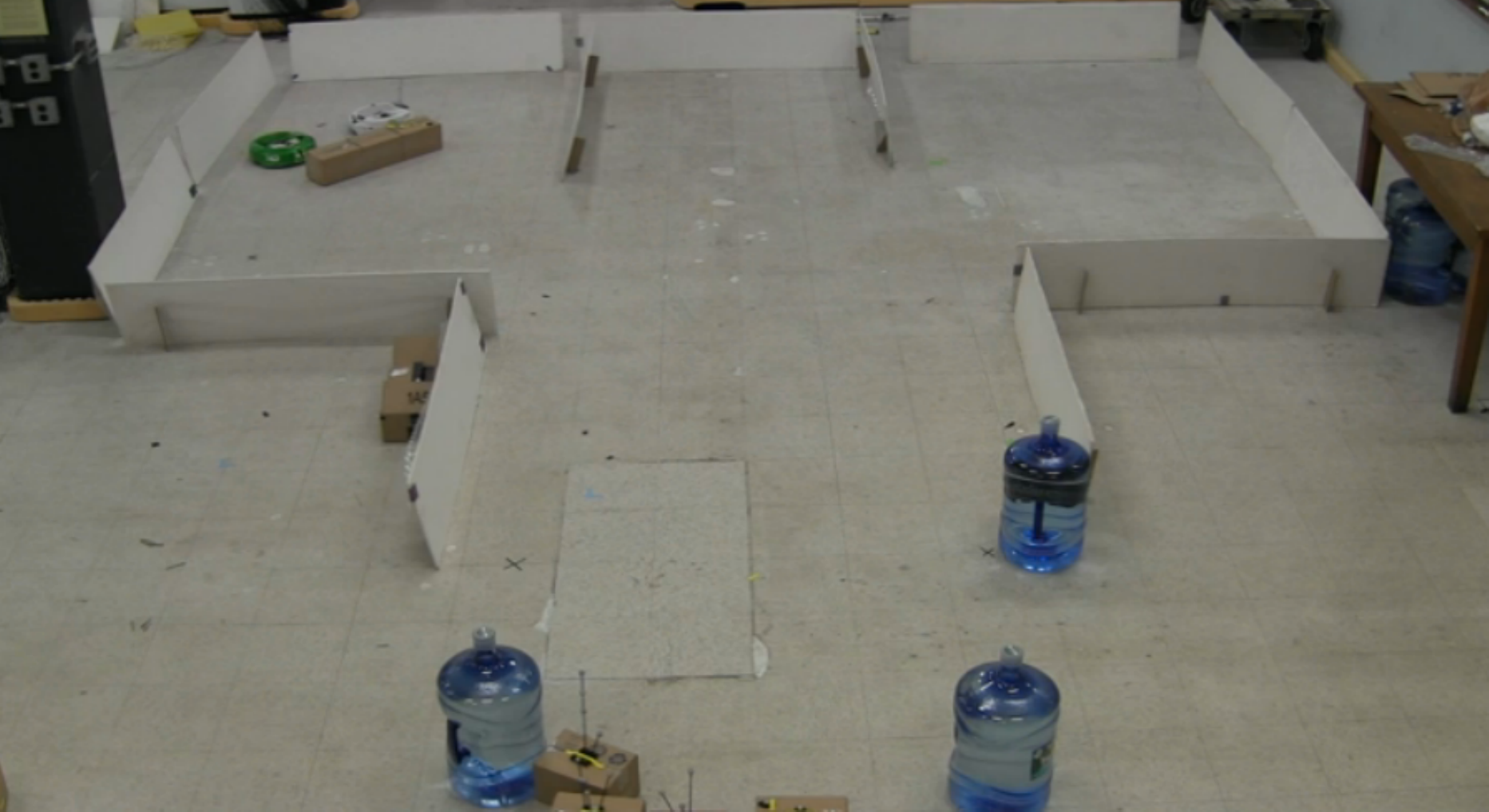}\label{Ng}}   \hfill 
\subfigure[The two robots push the large box back to the goal.]{\includegraphics[width=1.6in]{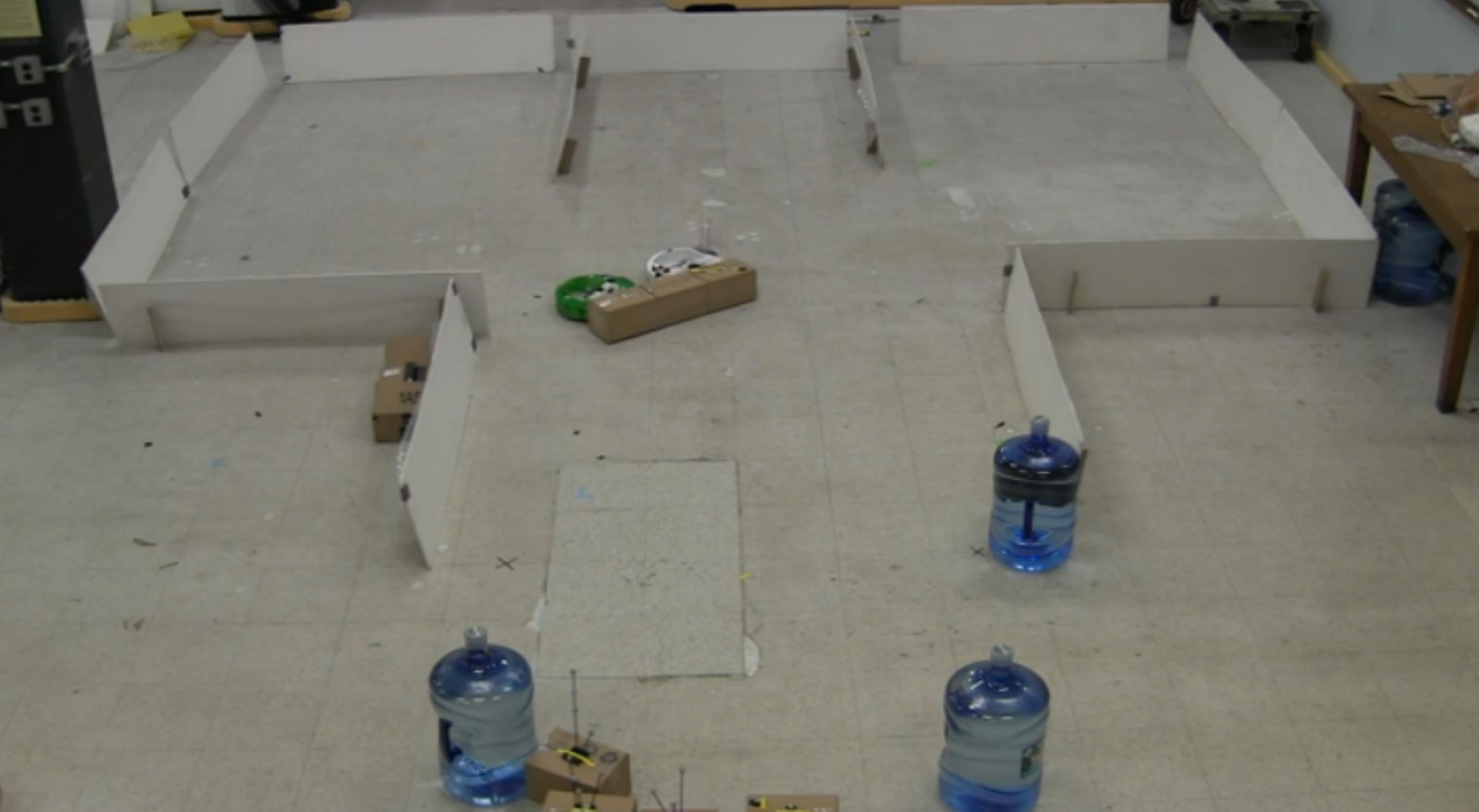}\label{Nh}}  
\caption{Video captures from the no communication version of the warehouse problem.}
\label{fig:NoCom} 
\end{figure*}

This case uses only two robots to more clearly show the result of not having communication as seen in Figure \ref{fig:NoCom}.
The policy generated by the planner assigns one robot to  go to each of the depots (Figure \ref{Na}). The robots then observe the contents of the depots they are in (Figure \ref{Nb}). If there are two robots in the same room as a large box, they will push it back to the goal. If there is only one robot in a depot and there is a small box to push, the robot will push the small box (Figure \ref{Nc}). 
If the robot is in a depot with a large box and no other robots, it will stay in the depot, waiting for another robot to come and help push the box (Figure \ref{Nd}). In this case, once the the other robot is finished pushing the small box (Figure \ref{Ne}), it goes back to the depots to check for other boxes or robots that need help (Figure \ref{Nf}). When it sees another robot and the large box in the depot on the left (depot 1), it attempts to help push the large box (Figure \ref{Ng}) and the two robots are successful pushing the large box to the goal (Figure \ref{Nh}).
In this case, the planner has generated a policy in a similar fashion to task allocation---two robots go to each room, and then search for help needed after pushing any available boxes.  However, in our case this behavior was generated by an optimization process that  considered the different costs of actions, the uncertainty involved and the results of those actions into the future. 


\subsection{Scenario 2: Local Communication}

The next scenario considered the case where robots could could communicate with each other when they were close to each other (approximately a meter). We added macro-actions to communicate and wait for communication. The macro-actions in this scenario were as follows:
\begin{itemize}
	\item Go to depot 1.
	\item Go to depot 2.
	\item Go to the drop-off area.
	\item Pick up the small box.
	\item Pick up the large box.
	\item Drop off a box.
	\item Go to an area between the depots (the ``waiting room").
	\item Wait in the waiting room for another robot.
	\item Send signal \#1.
	\item Send signal \#2.
\end{itemize}

Here, we allow the robots to choose to go to a ``waiting room" which is in between the two depots. This permits the robots to possibly communicate or receive communications before committing to going to one of the depots. The waiting room macro-action is applicable in any situation and terminates when the robot is between the waiting room walls. The depot macro-actions are not only applicable in the waiting room, while the drop-off, pick up and drop macro-actions remain the same. The wait macro-action is applicable in the waiting room and terminates when the robot observes another robot in the waiting room. The signaling macro-actions are applicable in the waiting room and are observable by other robots that are within approximately a meter of the signaling robot. 
However, note that \textit{we do not specify what sending each communication signal means}. 

\begin{figure*}
\subfigure[The three robots begin moving to the waiting room.]{\includegraphics[width=1.6in]{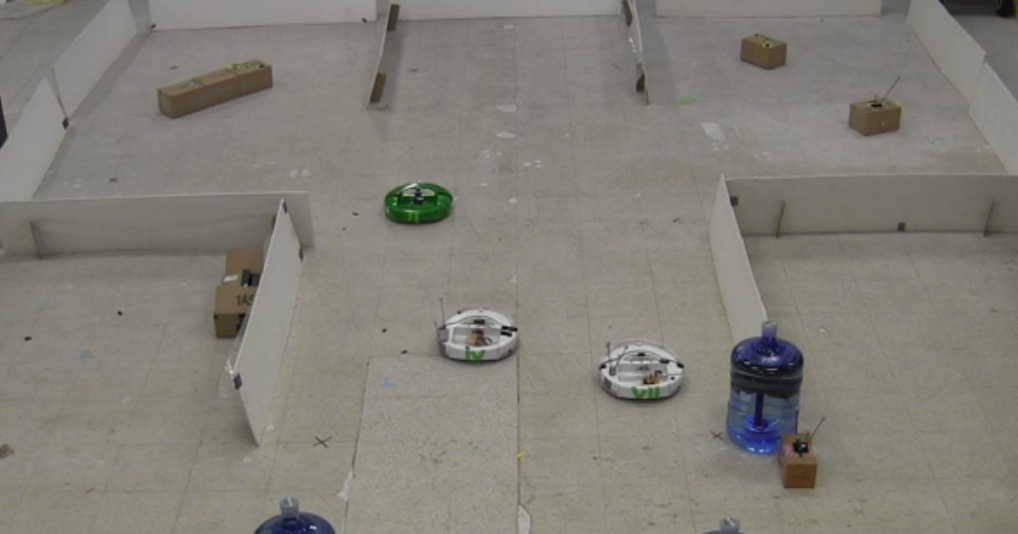}\label{La}} \hfill
\subfigure[One robot goes to depot 1 and two robots go to depot 2. The depot 1 robot sees a large box.]{\includegraphics[width=1.6in]{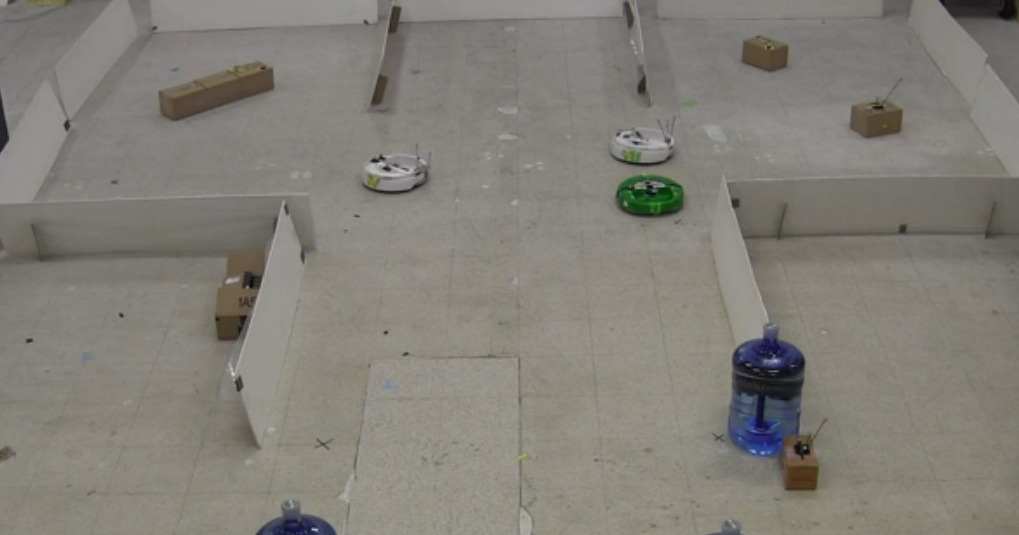}\label{Lb}}  \hfill
\subfigure[The depot 1 robot saw a large box, so it moved to the waiting room while the other robots pushed the small boxes.]{\includegraphics[width=1.6in]{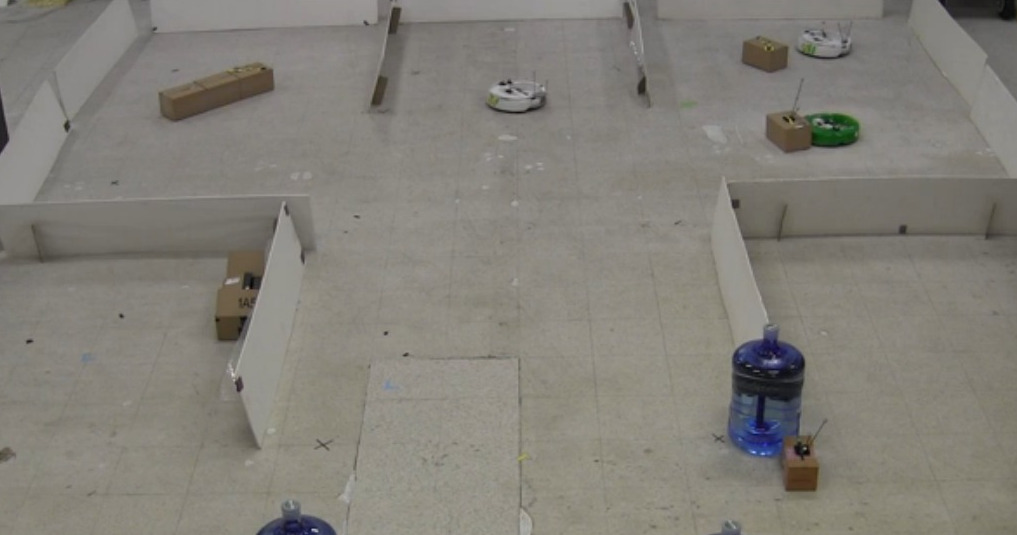}\label{Lc}}  \hfill
\subfigure[The depot 1 robot waits with the other robots push the small boxes.]{\includegraphics[width=1.6in]{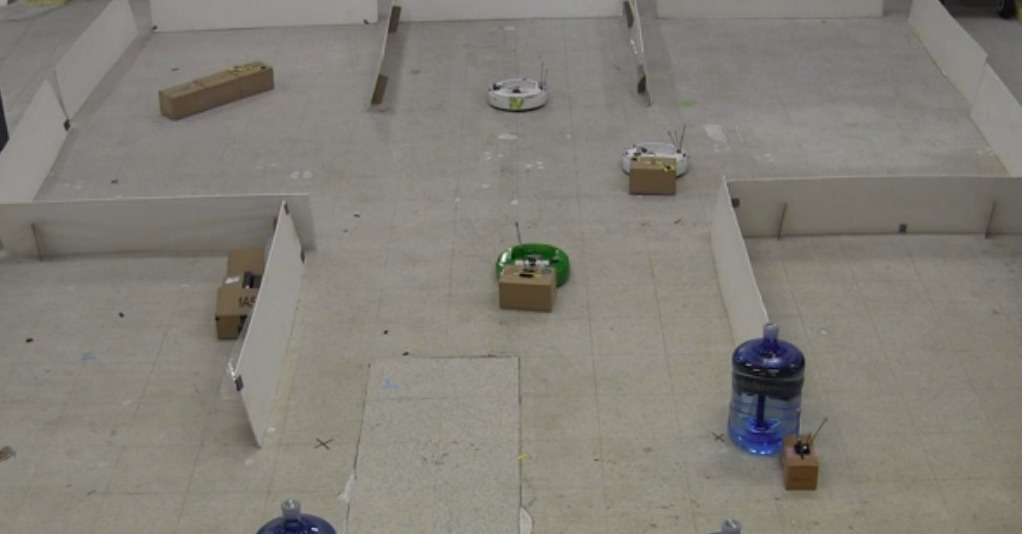}\label{Ld}}  \hfill
\subfigure[The two robots drop off the small boxes at the goal while the other robot waits.]{\includegraphics[width=1.65in]{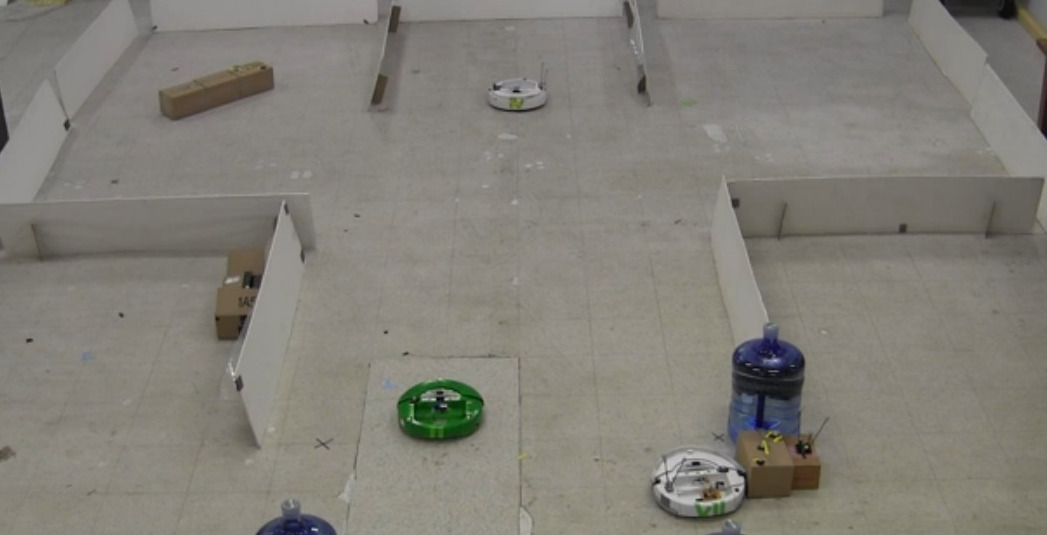}\label{Le}}  \hfill
\subfigure[The green robot goes to the waiting room to try to receive any signals. ]{\includegraphics[width=1.6in]{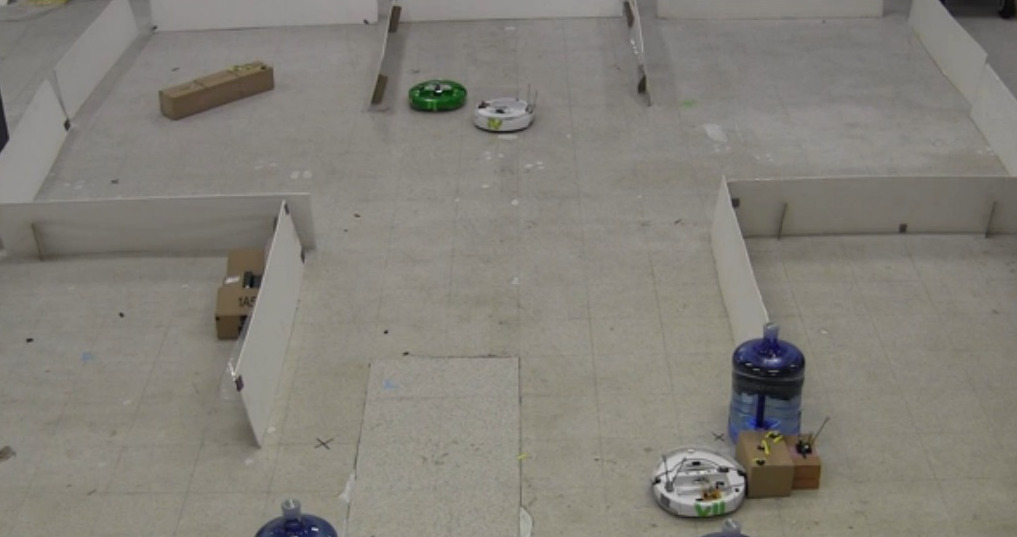}\label{Lf}}  \hfill
\subfigure[The white robot sent signal \#1 when it saw the green robot and this signal is interpreted as a need for help in depot 1.]{\includegraphics[width=1.59in]{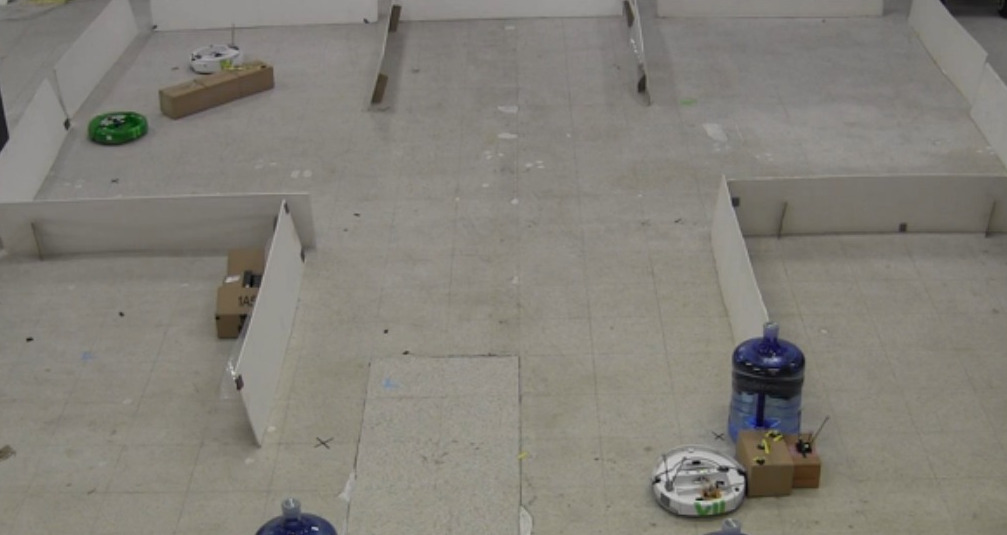}\label{Lg}}    \hfill 
\subfigure[The two robots in depot 1 push the large box back to the goal. ]{\includegraphics[width=1.6in]{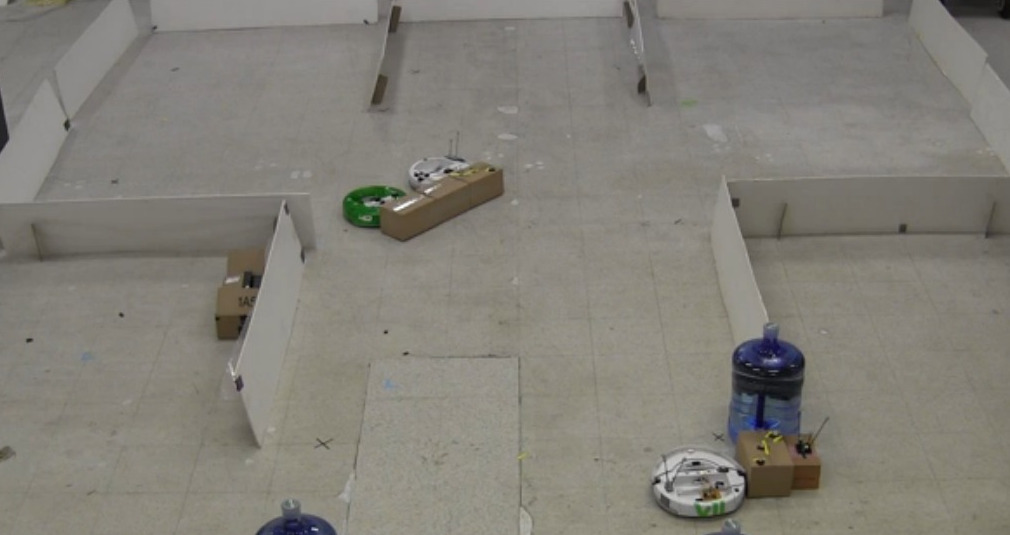}\label{Lh}} 
\caption{Video captures from the limited communication version of the warehouse problem.}
\label{fig:Limited} 
\end{figure*}


The results for this domain are shown in Figure \ref{fig:Limited}. We  see that the robots go to the waiting room (Figure \ref{La})
(because we required the robots to be in the waiting room before choosing to move to a depot) and then two of the robots go to depot 2 (the one on the right) and one robot goes to depot 1 (the one on the left) (Figure \ref{Lb}). Note that because there are three robots, the choice for the third robot is random while one robot will always be assigned to each of the depots. Because there is only a large box to push in depot 1, the robot in this depot goes back to the waiting room to try to find another robot to help it push the box (Figure \ref{Lc}). The robots in depot 2 see two small boxes and they choose to push these back to the goal (Figure \ref{Ld}). Once the small boxes are dropped off (Figure \ref{Le}), one of the robots returns to the waiting room (Figure \ref{Lf}) and then is recruited by the other robot to push the large box back to the goal (Figure \ref{Lg}). The robots then successfully push the large box back to the goal (Figure \ref{Lh}). Note that in
this case the planning process \textit{determines how the signals should be used to perform communication}.

\subsection{Scenario 3: Global Communication}

In the last scenario, we considered the case where the robots can use signaling (rather than direct communication) to request help. In this case, there is a light switch in each of the depots that can turn on a blue or red light. This light can be seen in the waiting room and there is another light switch in the waiting room that can turn off the light. (The light and switch were simulated in software and were not actually incorporated in the physical domain.)
As a result, the macro-actions in this scenario were as follows:
\begin{itemize}
	\item Go to depot 1.
	\item Go to depot 2.
	\item Go to the drop-off area.
	\item Pick up the small box.
	\item Pick up the large box.
	\item Drop off a box.
	\item Go to an area between the depots (the ``waiting room").
	\item Turn on a blue light.
	\item Turn on a red light.
	\item Turn off the light and go to depot 1.
	\item Turn off the light and go to depot 2.
\end{itemize}

Here, the robots can request help moving the large box by turning on a light. The first seven macro-actions are the same as for the communication case except we relaxed the assumption that the robots had to go to the waiting room before going to the depots (making both the depot and waiting room macro-actions applicable anywhere). The macro-actions for turning the lights on are applicable in the depots and the macro-actions for turning the lights off are applicable in the waiting room. 
Note that we did not assign a particular color to a particular depot, but instead allowed the planner to determine the best use of the light colors. 

\begin{figure}[t]
\subfigure[One robot starts first and goes to depot 1 while the other robots go to the waiting room.]{\includegraphics[width=1.6in]{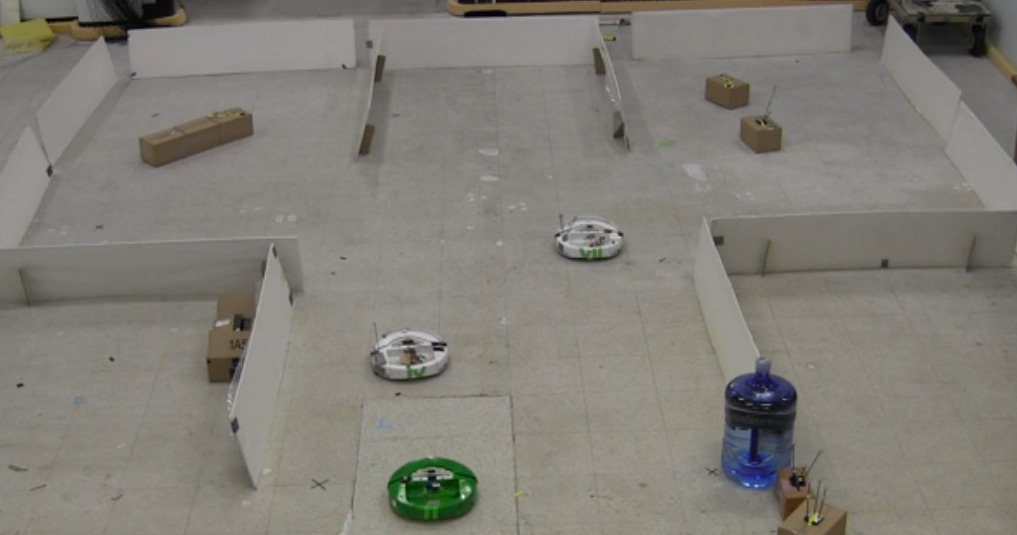}\label{Sa}} 
\subfigure[The robot in depot 1 sees a large box, so it turns on the red light (the light is not shown).]{\includegraphics[width=1.6in]{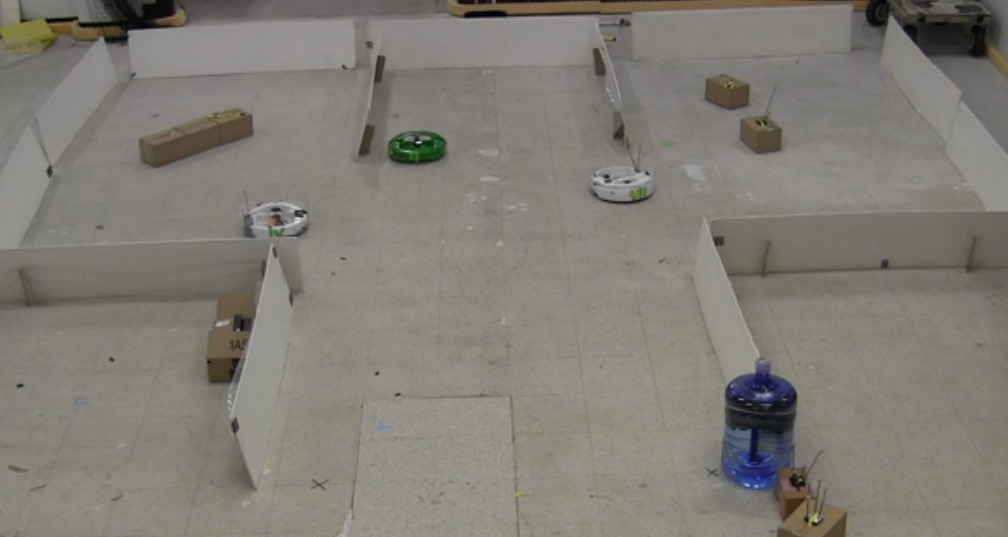}\label{Sb}} 
\subfigure[The green robot sees the light first, so it turns it off and goes to depot 1 while the white robot goes to depot 2.]{\includegraphics[width=1.6in]{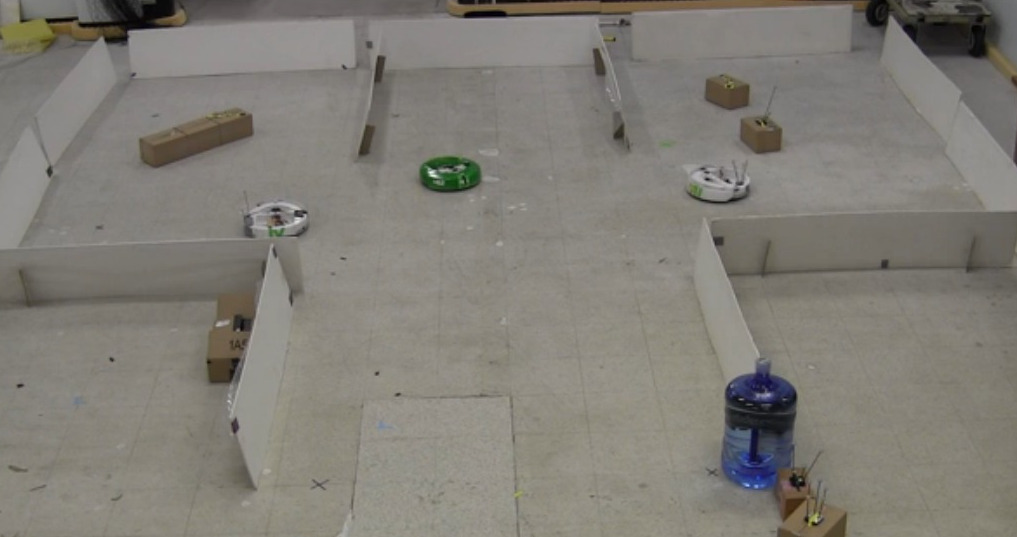}\label{Sc}} 
\subfigure[The robots in depot 1 move to the large box, while the robot in depot 2 begins pushing the small box.]{\includegraphics[width=1.6in]{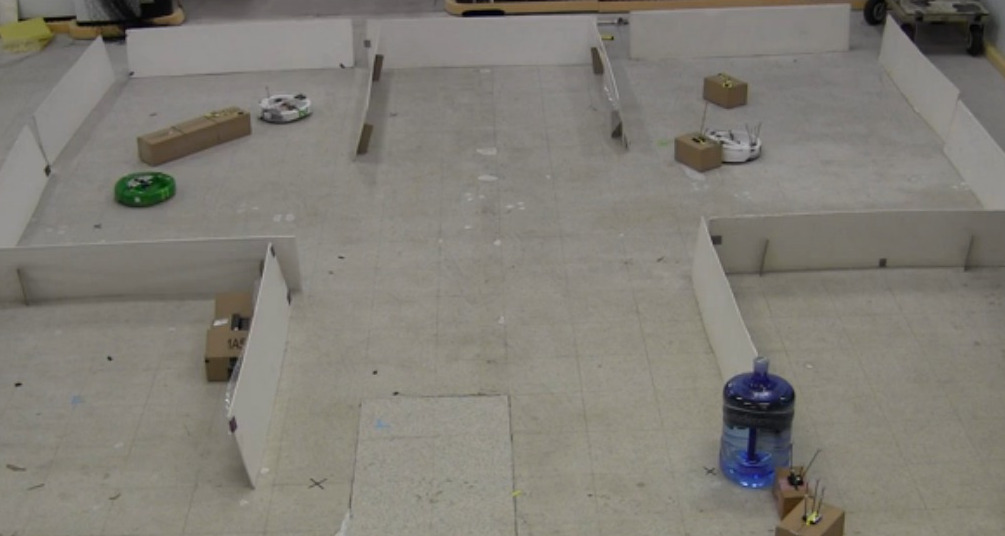}\label{Sd}} 
\subfigure[The robots in depot 1 begin pushing the large box and the robot in depot 2 pushes a small box to the goal. ]{\includegraphics[width=1.6in]{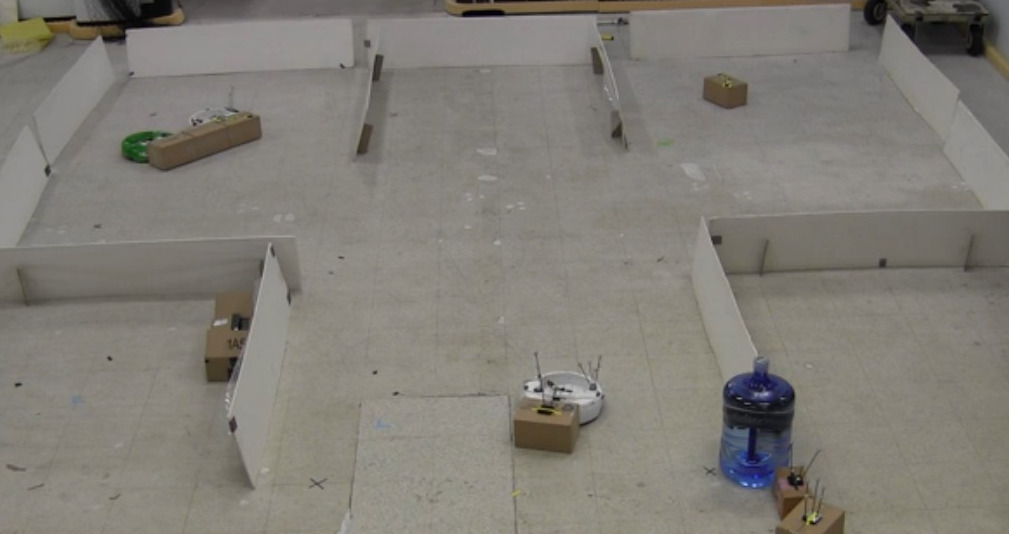}\label{Se}}  \hspace{1pt}
\subfigure[The robots from depot 1 successfully push the large box to the goal.]{\includegraphics[width=1.6in]{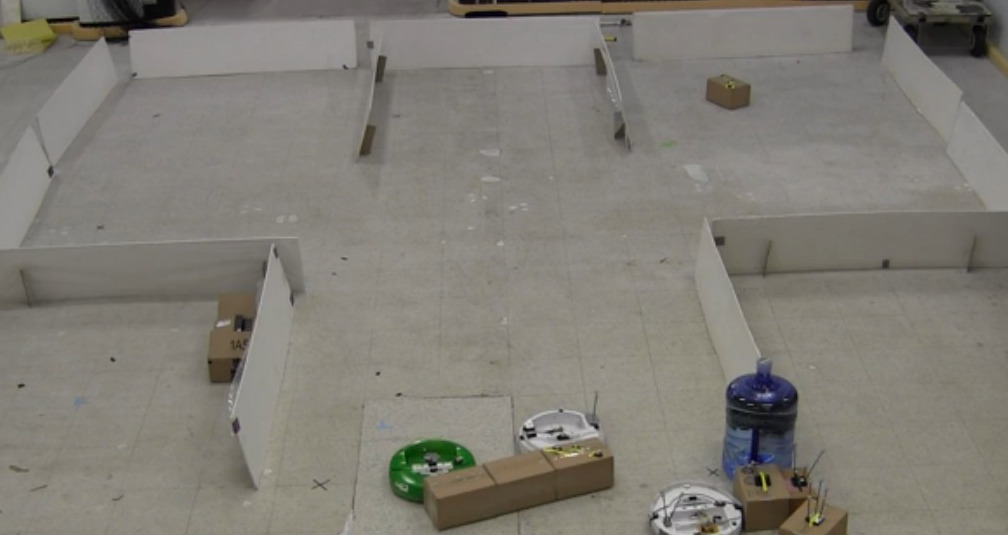}\label{Sf}} 
\caption{Video captures from the signaling version of the warehouse problem.}
\label{fig:Signal} 
\end{figure}

The results for this domain are shown in Figure \ref{fig:Signal}.  Because one robot started ahead of the others,  it was able to go to depot 1 to  sense the size of the boxes while the other robots go to the waiting room (Figure \ref{Sa}). 
 The robot in depot 1 turned on the light (red in this case, but not shown in the images) to signify that there is a large box and assistance is needed (Figure \ref{Sb}). The green robot is the first other robot to the waiting room, sees this light, interprets it as a need for help in depot 1, and turns off the light (Figure \ref{Sc}). The other robot arrives in the waiting room, does not observe a light on and moves to depot 2 (also Figure \ref{Sc}). 
 The robot in depot 2 chooses to push a small box back to the goal and the green robot moves to depot 1 to help the other robot (Figure \ref{Sd}). 
One robot then pushes the small box back to the goal while the two robots in depot 1 begin pushing the large box (Figure \ref{Se}). Finally, the two robots in depot 1 push the large box back to the goal (Figure \ref{Sf}).

\section{Related Work}

There are several frameworks that have been developed for multi-robot decision making in complex domains. For instance, behavioral methods have been studied for performing task allocation over time in loosely-coupled \cite{Parker98}  or tightly-coupled \cite{Stroupe04} tasks. These are heuristic in nature and make strong assumptions about the type of tasks that will be completed. 


One important related class of methods is based on using linear temporal logic (LTL) \cite{Belta07,loizou2004automatic}  to specify behavior for a robot;  from this specification, reactive controllers that are guaranteed to satisfy the specification can be derived.  These methods are appropriate when the world dynamics can be effectively described non-probabilistically and when there is a useful discrete characterization of the robot's desired behavior in terms of a set of discrete constraints.   When applied to multiple robots, it is necessary to give each robot its own behavior specification. 

Market-based approaches use traded value to establish an optimization framework for task allocation
\cite{Dias2003,Gerkey04}. These approaches have been used to solve real multi-robot problems \cite{Kalra2005},
but are largely aimed to tightly-coupled tasks, where the robots can communicate
through  a  bidding mechanism.

Emery-Montemerlo et al. \cite{Emery05} introduced a (cooperative) game-theoretic formalization of multi-robot systems which resulted in solving a Dec-POMDP. An approximate forward search algorithm was used to generate solutions, but because a (relatively) low-level Dec-POMDP was used scalability was limited. Also, Messias et al. \cite{Messias13} introduce an MDP-based model where a set of robots with 
controllers that can execute for varying amount of time must cooperate to solve a problem.
However, decision-making in their system is centralized.


\section{Conclusion} 
\label{sec:conclusion}

The MacDec-POMDP model is expressive enough to 
capture multi-robot systems of interest, but also simple enough to be feasible
to solve in practice.
Our results have shown that a general purpose MacDec-POMDP planner
can generate cooperative  behavior for complex multi-robot domains with task allocation, direct communication,
 and signaling behavior emerging automatically as properties
 of the optimal solution for the given problem model. 
 
The widespread use of techniques for solving much more restricted scenarios has led to a plethora
of usable algorithms for specific problems, but no way to combine these in more complex scenarios. 
In fact, our approach can build on the large amount of research in single and multi-robot systems that has gone into solving difficult problems such as navigation in a formation \cite{Balch98}, cooperative transport of an object \cite{Kube00}, coordination with signaling \cite{beckers1994local} or communication under various limitations \cite{rekleitis2004limited}. Many of the solutions for these problems may be able to be represented as macro-actions in our framework, bootstrapping state-of-the-art research to solve even more complex multi-robot problems. 

In the future, we plan to explore incorporating these state-of-the-art macro-actions into our MacDec-POMDP framework as well as examine other types of structure that can be exploited. 
Other topics we plan to explore include increasing scalability by making solution complexity depend on the number of agent interactions rather than the domain size, and having robots
learn models of their sensors, dynamics and other robots. 
These approaches have great potential to lead to
automated solution methods for general multi-robot coordination problems with large numbers of heterogeneous robots in complex, uncertain domains.



\bibliographystyle{aaai}
\bibliography{OptBib}

\end{document}